\documentclass[utf8]{FrontiersinHarvard}

\usepackage{algorithm}%
\usepackage{algorithmicx}%
\usepackage{algpseudocode}%
\usepackage{listings}%
\definecolor{azul}{rgb}{0.1,0.2,0.7}
\definecolor{verde}{rgb}{0.4,0.6,0.4}
\definecolor{bordo}{rgb}{0.8,0.3,0.3}
\definecolor{rojo}{rgb}{0.8,0.3,0.3}
\definecolor{gris}{rgb}{0.4,0.4,0.4}
\definecolor{amarillon}{RGB}{245,226,190}
\definecolor{amarillon1}{RGB}{253,224,180}
\definecolor{amarillon2}{RGB}{247,250,210}
\definecolor{naranja}{RGB}{255,150,102} 
\usepackage{graphicx}               % Necessary to use \scalebox
\usepackage{color,xcolor}
\usepackage{verbatim}
\usepackage{lipsum}
\usepackage{slashed}
\usepackage{ragged2e}
\usepackage[nomath,light]{kpfonts} % tipografía kpfonts 
\usepackage{yfonts} %Frak Tipo
\usepackage{eufrak}
\usepackage{upgreek}
\usepackage{url}
\usepackage[export]{adjustbox}
\usepackage{mathrsfs}   % cambio de tipografía cursiva
\usepackage[scr=dutchcal,calscaled=1]{mathalfa}
\usepackage{relsize} % bloques de tamaño variable
\expandafter\let\csname equation*\endcsname\relax
\expandafter\let\csname endequation*\endcsname\relax

\usepackage{amsmath}
\usepackage{amsthm}

\theoremstyle{remark}      % estilo para observaciones, notas, etc.
 
\usepackage{amssymb}
\usepackage{stmaryrd} %algunso símbolos e.g. \llbracket 
\usepackage{tikz}
\usepackage{tikz-cd}
\usetikzlibrary{automata}
\usetikzlibrary{topaths}
\usepackage{pstricks}
\usetikzlibrary{automata, positioning, calc}
\usetikzlibrary{arrows}
\usetikzlibrary{mindmap, trees}

\def\keyFont{\fontsize{8}{11}\helveticabold }
\def\firstAuthorLast{Caruso {et~al.}} %use et al only if is more than 1 author
\def\Authors{Mariano Caruso\,$^{1,2}$ and Cecilia Jarne\,$^{3,4,5,6,*}$}
\def\Address{$^{1}$Universidad Internacional de La Rioja (UNIR), La Rioja, Spain.

$^{2}$FIDESOL, Granada, Spain.

$^{3}$Departamento de Ciencia y Tecnolog\'ia, Universidad Nacional de Quilmes (UNQ), Bernal, Buenos Aires, Argentina.

$^{4}$CONICET, Buenos Aires, Argentina.

$^{5}$Center of Functionally Integrative Neuroscience (CFIN), Department of Clinical Medicine, Aarhus University, Aarhus, Denmark.

$^{6}$ Latin American Brain Health Institute (BrainLat), Universidad Adolfo Ibàñez.}

\def\corrAuthor{Cecilia Jarne}
\def\corrEmail{cecilia.jarne@unq.edu.ar}
\usepackage[normalem]{ulem}
\begin{document}
\onecolumn
\firstpage{1}

\title {Analysing Rescaling, Discretisation, and Linearisation in RNNs for Neural System Modelling} 

\author[\firstAuthorLast ]{\Authors} %This field will be automatically populated
\address{} %This field will be automatically populated
\correspondance{} %This field will be automatically populated

\extraAuth{}% If there are more than 1 corresponding author, comment this line and uncomment the next one.
%\extraAuth{corresponding Author2 \\ Laboratory X2, Institute X2, Department X2, Organization X2, Street X2, City X2 , State XX2 (only USA, Canada and Australia), Zip Code2, X2 Country X2, email2@uni2.edu}

\maketitle

\begin{abstract}

Recurrent Neural Networks (RNNs) are widely used to model neural activity in Computational Neuroscience. Here, we explore the mathematical foundations of three fundamental procedures {that can be} implemented: temporal rescaling, discretisation, and linearisation. These techniques provide crucial tools for characterising the behaviour of RNNs, offering insights into their temporal dynamics, facilitating practical computational implementation, and allowing for linear approximations for analysis. We discuss the flexible order in which these procedures can be applied, emphasising their importance in modelling and analysing RNNs for neuroscience and {formally prove that these three operations commute pairwise}. We also explicitly describe the conditions under which these procedures can be considered interchangeable. Our findings directly inform the design of biologically plausible $\mathtt{RNN}$ models for simulating neural dynamics observed in decision-making circuits and motor control, where temporal scaling and stability are critical for matching experimental recordings. {Furthermore, we show that this exact commutativity guarantees the structural preservation of the network's controllability, preventing the emergence of inaccessible state-spaces under numerical discretisation or temporal rescaling.}

\keyFont{ \section{Keywords:} RNNs, rescaling, discretisation, linearisation, computational neuroscience}
\end{abstract}

\section{INTRODUCTION}

$\mathtt{RNN}$s are universal approximators of dynamical systems (\cite{10.1007/11840817_66, FUNAHASHI1993801}). They have become invaluable tools in computational neuroscience over the last 40 years (\cite{Hopfield3088, DBLP:journals/neco/SussilloB13, Pandarinath9390}). Networks typically represent certain cortical areas from an abstract set of recurrently connected units (see left panel of Figure 1). They are particularly effective in capturing the time-varying dynamics of neural processes, allowing us to simulate and analyse complex brain functions and interactions, such as motor control, decision-making and other complex processes (\cite{RUSSO2020745, Bi10530, annurev-neuro-091724-015512, Jarne_2021}). Their ability to capture sequential dependencies and recurrent patterns makes them well-suited for tasks such as understanding information processing in the brain and decoding neural signals. $\mathtt{RNN}$s also have various applications in machine learning, including sequential data analysis and time-series prediction (\cite{ZHANG2023143}), and are also used to process brain data.

{There are three main procedures we can consider applying to differential equations, particularly in the context of neuroscience and recurrent neural networks ($\mathtt{RNN}$s). These procedures $-$ temporal rescaling, temporal discretisation, and linearisation $-$ are essential for characterising the behaviour of the systems represented by these models. We will discuss the outcomes of applying these procedures in different sequences and examine the conditions under which the order of application does not affect the results. In other words, we will explore when these procedures commute within the $\mathtt{RNN}$ framework. Additionally, we provide a brief introduction to each of these procedures. They are directly related to the design of biologically plausible $\mathtt{RNN}$ models used for simulating the neural dynamics observed in decision-making circuits (\cite{WANG2002955}) and motor control systems (\cite{Churchland2012}), where temporal scaling and stability are critical for aligning with experimental recordings.} {While each of these operations is individually well understood, their pairwise commutativity within the $\mathtt{RNN}$ framework used in computational neuroscience has not, to our knowledge, been formally established in the literature. The present work fills this gap by providing explicit proofs and a unified diagrammatic representation.}

Temporal rescaling can improve the numerical stability of solving differential equations. By rescaling the time variable, one can potentially reduce the condition number of the underlying linear system, which can result in more accurate and stable numerical solutions, especially when using numerical integration methods. This approach is especially useful in scenarios such as simulating long-term dynamics or processes that occur over a wide range of timescales. Temporal rescaling allows us to efficiently capture the behaviour of a system. By adjusting the time units, we can focus computational resources where they are most needed, avoiding unnecessary computations at very short or very long times. It can help identify dominant modes, equilibrium points, or oscillations in the system and gain a better understanding of the underlying dynamics.

Transitioning from a continuous-time $\mathtt{RNN}$ to one suitable for computer implementation involves discretisation. In continuous-time $\mathtt{RNN}$s, dynamics is modelled using differential equations, which describe how neuron activations change continuously over time. However, computers operate in discrete time, meaning that they process information in discrete time steps. This is well known, and we daily employ numerical methods to approximate continuous-time behaviour in a discrete-time framework. Given a small time step, often denoted as $\Delta$, to divide time into discrete intervals. Then, we use the Euler method, or more advanced techniques to iteratively update the neuron activations at each time step (\cite{cheney2018numerical}). {Although a variety of numerical schemes can be used to discretise continuous-time RNNs, the present analysis focuses on the forward Euler method. This choice is not merely computational but also structural. The exact commutativity results established in this work rely on the fact that Euler discretisation evaluates the vector field only once per time step, yielding an affine update that preserves the algebraic independence between discretisation, temporal rescaling, and local linearisation. In contrast, higher-order schemes such as Runge–Kutta methods construct the update rule through multiple intermediate evaluations of the nonlinear vector field. These nested nonlinear compositions introduce higher-order interaction terms that generally prevent the resulting discrete map from being identical to the one obtained by first linearising the continuous system and subsequently discretising it. Consequently, the exact commutativity proven here should be understood as a property of the Euler framework and not as a generic feature of arbitrary numerical integrators.}

{From a practical perspective, this restriction is also consistent with common RNN implementations in computational neuroscience and machine learning. Higher-order integrators increase the number of function evaluations required at each update step and therefore substantially raise both training and simulation costs, particularly when gradients must be propagated through long recurrent trajectories. Similar trade-offs have been discussed in the Neural ODE literature, where increased solver accuracy is often accompanied by significantly higher computational demands (\cite{chen2018neural}). Furthermore, the Euler formulation preserves a direct correspondence between the continuous-time firing-rate equation and its discrete implementation, maintaining a transparent interpretation in terms of passive decay and recurrent synaptic input. Investigating whether weaker forms of commutativity survive under higher-order or adaptive integration schemes remains an interesting direction for future work.}

The key idea is to approximate continuous differential equations, such as those governing the $\mathtt{RNN}$'s dynamics, using finite differences. This process effectively transforms continuous-time $\mathtt{RNN}$ equations into a discrete-time form. {This transition allows us to use computers for training and inference while still capturing essential aspects of the continuous-time model's behaviour, enabling the application of $\mathtt{RNN}$s to develop computational models of the brain and cognitive tasks.} It is important to note that despite this discretisation, the behaviour of the dynamical system can be characterised. The discrete-time $\mathtt{RNN}$'s stability, for instance, can be assessed by examining the eigenvalues of its weight matrix, shedding light on the presence and stability of fixed points in the network dynamics (\cite{DBLP:journals/neco/SussilloB13, Jarne2023}). 

Linearisation of dynamical systems simplifies complex systems into linear approximations, making it (sometimes) easier to analyse and understand their behaviour. Linear systems are well-studied and have well-established mathematical tools for their analysis. {Linearisation helps to identify stable modes in cortical networks (\cite{DBLP:journals/neco/SussilloB13})}. Effects of different linearisation mechanisms in $\mathtt{RNN}$s have been discussed before (\cite{pagan2023brief}). The authors compare the $\mathtt{RNN}$ dynamics that can be written in terms of the "activations" or "activities", meaning that $\mathtt{RNN}$'s dynamics can be written in terms of the net inputs to each unit before the pointwise nonlinearity or in terms of the output of each unit after the pointwise nonlinearity.

We studied the interrelation of these three procedures: rescaling, discretisation, and linearisation, commonly used to create frameworks and explore $\mathtt{RNN}$s and the performance, dynamics, and mechanisms when they are trained for different tasks.

{The chosen assumptions and simplifications have direct consequences on the models and predictions. For example, in (\cite{Wang2018}), the authors considered a variant of the interval production task termed the cue-set-go (CSG) task and studied such in real data and trained $\mathtt{RNN}$s. Neurons recorded during the CSG task display nonlinear, nonmonotonic activity that is temporally compressed on short interval trials and stretched on long interval trials, a phenomenon that has been termed temporal scaling. When all the neurons are temporally scaled by a certain factor without changing the response profile, the population activity goes through the same continuum of states, but only at a different speed. Therefore, in the state space, temporal scaling manifests as similar neural trajectories that evolve at different speeds. { In that study, a linearisation of the $\mathtt{RNN}$ dynamics was employed to investigate how temporal scaling arises from changes in input; the commutativity between rescaling and linearisation formalised here provides a rigorous foundation for such approaches.}

\section{Methods}

Consider a collection of $N$ artificial neurons, each associated with a dynamic quantity termed activity, which is described by a function $h_i: [a, b] \longrightarrow \mathscr{H} \subseteq \mathbb{R}$ for $i=1, \cdots, N$. These $N$ functions can be organized into a column vector $\pmb{h} {=} (h_1, \cdots, h_i, \cdots, h_N)^{T}$, commonly referred to as the \textit{hidden} layer of the $\mathtt{RNN}$, where $T$ indicates the transpose operation. The vector $\pmb{h}$ encapsulates the network's activity state at time $t$, encompassing all $N$ neurons. Additionally, there are $M$ input functions, $x_k: [a, b] \longrightarrow X \subseteq \mathbb{R}$ for $k=1, \cdots ,M$, which can be similarly arranged into a column vector $\pmb{x} {=} (x_1, \cdots, x_i, \cdots, x_M)^{T}$. For recurrent neural networks, the activity vector $\pmb{h}$ follows the differential equation:

\begin{equation}\label{din red}
\pmb{h}'(t)=-\lambda\,\pmb{h}(t) + \pmb{\sigma}\pmb{(}\pmb{w\,h}(t)+\widetilde{\pmb{w}}\,\pmb{x}(t)\pmb{)}.
\end{equation}
\label{eq-01}

Here, $\pmb{h}'(t)$ denotes the time derivative in the standard sense. The matrices $\pmb{w}$ and $\widetilde{\pmb{w}}$ have dimensions $N\times N$ and $N\times M$, respectively, with the elements $w_{ij}$ of $\pmb{w}$ representing synaptic connections, similarly for $\widetilde{\pmb{w}}$. The activation vector field $\pmb{\sigma}$ maps $\mathbb{R}^N$ to itself, and satisfies $\pmb{\sigma}(\pmb{0}){=}\pmb{0}$, reflecting the principle that neuronal activity cannot spontaneously regenerate; in other words, activating a neuron with zero initial activity yields zero output. Each component of $\pmb{\sigma}$ is derived by applying a distinct non-linear function $\sigma:\mathbb{R}{\longrightarrow}\mathbb{R}$. For a vector $\pmb{\varphi}\in \mathbb{R}^N$, expressed as $\pmb{\varphi}=(\varphi_1,\cdots,\varphi_N)$, the vector $\pmb{\sigma}(\pmb{\varphi})$ is given by $(\sigma(\varphi_1),\cdots,\sigma(\varphi_N))$. Furthermore, $\lambda^{-1}$ represents the decay time of each signal $h_i$ when the network is entirely disconnected, meaning $\pmb{w}=\pmb{0}$ and $\widetilde{\pmb{w}}=\pmb{0}$.

The network's activity state is determined by \eqref{din red}, which is updated as a result of the interaction between them via $\pmb{w}$, with external signals $\pmb{x}$ influencing the neurons' activity according to $\widetilde{\pmb{w}}$, along with some initial condition. Instead of starting from \eqref{din red}, it may be useful to start from the "biased" version of the equation as follows

\begin{equation}\label{din red bias}
\pmb{h}'(t)=-\lambda\,\pmb{h}(t) + \pmb{\sigma}\pmb{(}\pmb{w\,h}(t)+\pmb{b}+\widetilde{\pmb{w}}\,\pmb{x}(t)\pmb{)}.
%\pmb{h}'(t)=-\frac{\,1\,}{\tau}\pmb{h}(t) + \pmb{\sigma}\pmb{(}\pmb{w\,h}(t)+\pmb{b}+\widetilde{\pmb{w}}\,\pmb{x}(t)\pmb{)}
\end{equation}
\label{eq-02}

In Equation \ref{eq-02}, $\pmb{b}=(b_1,\cdots,b_i,\cdots,b_N)^T$ is another column vector, and each component is the bias for each activity $h_i$. We can write \eqref{din red} in terms of its components 

\begin{equation}\label{din red comp}
h_i'(t)=-\lambda\,h_i(t) + \sigma\Big{(}\sum\nolimits_{j=1}^N w_{ij}h_j(t)+b_i+\sum\nolimits_{k=1}^M \widetilde{w}_{ik}x_k(t)\Big{)}.
%h_i'(t)=-\frac{\,1\,}{\tau}h_i(t) + \sigma\Big{(}\sum\nolimits_{j=1}^N w_{ij}h_j(t)+b_i+\sum\nolimits_{k=1}^M \widetilde{w}_{ik}x_k(t)\Big{)}.
\end{equation}

To better understand, we will work with a more compact notation:
\begin{equation}\label{din general}
\pmb{h}'(t)=\pmb{F}\pmb{(}\pmb{h}(t),\pmb{x}(t)\pmb{)}.
\end{equation} 

We use this compact notation 
$\pmb{F}\pmb{(}\pmb{h}(t),\pmb{x}(t)\pmb{)}{=}-\lambda\,\pmb{h}(t) + \pmb{\sigma}\pmb{(}\pmb{w\,h}(t)+\pmb{b}+\widetilde{\pmb{w}}\,\pmb{x}(t)\pmb{)}$ and it will be useful to consider the matrices $\pmb{A}:=\pmb{w}-\lambda\pmb{I}$ and $\pmb{B}:=\widetilde{\pmb{w}}$, where $\pmb{I}$ is the $N\times N$ identity matrix. 

Although we don't see any "recurrence" relationship in the discrete mathematics sense, {the germ of this concept is already present in} \eqref{din general}, indicating that changes in $\pmb{h}$ over time depend on its current state.

As previously mentioned, we considered three main procedures that can be applied to this differential equation: temporal rescaling, temporal discretisation, and linearisation. We categorise the first two as different forms of rescaling, while we will address the discretisation process separately. We will demonstrate that the outcomes of applying these procedures are independent of the order in which they are applied. In other words, we will prove that these procedures are mutually commutative under certain conditions.

\subsection{Rescaling}\label{rescaling}

Informally, time rescaling operator takes the activity function \(\pmb{h}(t)\) and \textit{redraws} it on a new time scale. Instead of analysing \(\pmb{h}\) as a function of time \(t\), we define a new variable \(s\) such that \(t = \tau s\). If \(\tau >1\), the timeline is stretched (events occur later), and if \(0<\tau < 1\) the timeline is compressed (events occur more rapidly). Formally, the time rescaling function of parameter $\tau > 0$ is given by $\varrho_\tau{:}\,s\longmapsto t=\tau s$. We introduce the rescaling operation by a factor $\tau$ for the activity $\pmb{h}(t)$ as 

\begin{equation}
\mathscr{R}_\tau\pmb{\big(}\pmb{h}(t)\pmb{\big)}(s){=}(\pmb{h}\circ\varrho_\tau)(s).
\label{eq-04}
\end{equation}

We can apply the temporal rescaling operation ($\mathscr{R}_\tau$) on the network's activity vector $\pmb{\mathfrak{h}}{=}\mathscr{R}_\tau\pmb{(h)}$ given by the Equation \ref{din general}. The external excitation vector $\pmb{\chi}{=}
\mathscr{R}_\tau\pmb{(x)}$, and applying the chain rule for the time derivative $\pmb{h}'$, then $\pmb{\mathfrak{h}}'(s)=\tau\pmb{F}\pmb{(}\pmb{\mathfrak{h}}(s),\pmb{\chi}(s)\pmb{)}$. Finally, under rescaling operation  $\mathscr{R}_\tau$, the expression  \eqref{din red bias} take the form

\begin{equation}\label{eq:second_equation}
\pmb{\mathfrak{h}}'(s)=-\tau\lambda\,\pmb{\mathfrak{h}}(s)+\tau\,\pmb{\sigma}\left(\pmb{\omega} \cdot \pmb{\mathfrak{h}}(s)+ \widetilde{\pmb{\omega}} \cdot\pmb{\chi}(s)+\pmb{b}\right).
\end{equation}

Comparing this expression with \eqref{din red bias}, $\tau$ reappears in front of $\pmb{\mathfrak{h}}$ and $\pmb{\sigma}$. We can see how the temporal rescaling affects the characteristic relaxation time of the network in the absence of interaction $\lambda^{-1}\longmapsto(\tau\lambda)^{-1}$ and also the activation function. The maximum and minimum amplitude of the new activation function is modified by this operation.
{Every scaling transformation in the time domain has its counterpart in the frequency domain. Let \(\pmb{H}(\omega)\) be the Fourier transform of \(\pmb{h}(t)\), then \(\tau^{-1} \pmb{H}(\tau^{-1} \omega)\) will be the Fourier transform of {\(\mathscr{R}_\tau\pmb{(}\pmb{h}(t)\pmb{)}\)}. This result can be compactly expressed in operational terms as $\mathscr{F} \circ \mathscr{R}_\tau = \tau^{-1} \mathscr{R}_{\tau^{-1}} \circ \mathscr{F}$, where \(\mathscr{F}\) denotes the Fourier transform, so that \(\pmb{H}(\omega) = \mathscr{F}[\pmb{h}(t)](\omega)\). It is noteworthy that a temporal rescaling in the model also affects the amplitude of the spectral activity.}

\subsection{Discretisation}\label{discrete}

To compute \eqref{din general}, in the sense of using a computer to perform simulations, it is always necessary to perform a discretisation process. Let us consider the time interval to study such system, $[a, b] \subset \mathbb{R}$, we can apply a well-known procedure, which consists of cutting $n$ pieces of equal size of $\Delta=(b-a)/n$, to obtain a sequence $\{t_0,t_1,\cdots,t_n\}$, where $t_0=a$, and $t_{n}=b$. We introduce the discretisation operation associated with the sequence of times $t_k=a+ k\Delta, \,k=0,\cdots,n$,  for the activity $\pmb{h}(t)$ as

\begin{equation}\label{discretisation operation}
\mathscr{D}_\Delta\pmb{\big(}\pmb{h}(t)\pmb{\big)}{=}\pmb{h}(t_k).
\end{equation}

From the slicing $\{t_k\}_k$, the sequence of snapshots of corresponding neural activities $\{\pmb{h}(t_k)\}_k$ is defined. 
The operation \eqref{discretisation operation} can be applied also on the external excitation $\mathscr{D}_\Delta\pmb{\big(}\pmb{x}(t)\pmb{\big)}{=}\pmb{x}(t_k)$. For very small $\Delta$ 

\begin{equation}
\pmb{h}'(t_k)\simeq \big{[} \pmb{h}(t_k+\Delta)-\pmb{h}(t_k) \big{]}/\Delta,
\end{equation}

from the sequence of times, we have $t_{k+1}=t_k+\Delta$, valid for $k=0,\cdots, n-1$. {By definition,}

\begin{equation}\label{derivada milaneseada}
\mathscr{D}_\Delta\pmb{\big(}\pmb{h}'(t)\pmb{\big)}=\pmb{h}'(t_k)%=\big{[} \pmb{h}(t_{k+1})-\pmb{h}(t_k) \big{]}/\Delta. 
\end{equation}

So, under \eqref{derivada milaneseada}, now considered as an exact equality, the differential equation \eqref{din general} can be rewritten as $\pmb{h}(t_{k+1})=\pmb{h}(t_{k})+\pmb{F}\pmb{(}\pmb{h}(t_k),\pmb{x}(t_k)\pmb{)}\Delta$. Finally, applying the discretisation operator $\mathscr{D}_\Delta$ on the expression \eqref{din red bias} takes the form 

\begin{equation}\label{din general milaneseada}
\pmb{h}(t_{k+1})=\pmb{h}(t_{k})-\lambda\,\pmb{h}(t_k) {\cdot}\Delta + \pmb{\sigma}\pmb{(}\pmb{w\,h}(t_k)+\pmb{b}+\widetilde{\pmb{w}}\,\pmb{x}(t_k)\pmb{)}{\cdot}\Delta
\end{equation}

This serves as a recurrence relation; given the information of activity and excitation signals in a given slice, let's say at time $t_k$, this relation gives us the value of the activity signal for the next slice at time $t_{k+1}$. {This form is well-suited for numerical computation, but its direct evaluation by hand is impractical owing to the repetitive nature of the iteration, particularly when the map $\pmb{F}$ contains nonlinear terms}. 

In Computational Neuroscience, the fixed points of the $\mathtt{RNN}$ models defined by equations \eqref{din red} and \eqref{din red bias} are frequently used to simulate neural responses to static or slowly changing stimuli. Such equations are more commonly used explicitly in Computational Neuroscience, while equation \eqref{din general milaneseada} is more common in machine learning, but they are closely related. Equation \eqref{din general milaneseada} has the same fixed point as equation \eqref{din red}. Still, hyperbolic stability is obtained when eigenvalues have a magnitude less than 1. {It should be noted that this eigenvalue condition strictly applies when the continuous-time system has already been linearised; for the full nonlinear system, the correspondence between the magnitude of the discrete map's eigenvalues and asymptotic stability is only guaranteed locally, in a neighbourhood of a fixed point.} Hence, if a fixed point is stable for equation \eqref{din general milaneseada}, it is also stable for equation \eqref{din red comp}, but the converse is not true (\cite{zhu2023learning}).In computational neuroscience, $\mathtt{RNN}$s of the form of equation \eqref{din red} are studied using such mapping and also for their fixed point properties (\cite{DBLP:journals/neco/SussilloB13}). 

{Additionally, the choice of $\Delta$ in the discretisation can directly affect the model's ability to capture phenomena such as gamma oscillations (30-80 Hz) in neuronal activity, critical in attention processes (\cite{annurev-neuro-062111-150444}).}

\subsection{Linearisation}\label{linear}

 {Linear approximations around fixed points are widely used to study attractor dynamics in cortical networks (\cite{Brunel2001}). Our framework formalises conditions under which such approximations hold, which is critical for interpreting stability in models of persistent activity.}

Since the non-linear object in $\pmb{F}$ is exclusively in $\pmb{\sigma}$, linearisation is a procedure related to the \textit{activation field} and the type of signals (neuronal activity) we are interested in considering.

Within this activation field $\pmb{\sigma}$, let's now examine each function within $\sigma:\mathbb{R}\longrightarrow \mathbb{R}$ and assume that $\sigma(\varphi)$ is $k$-times differentiable at $\varphi=0$. By Taylor's theorem, there exists a \textit{remainder} function $R_k(\varphi)$ that allows us to write it as

\begin{equation}
\sigma(\varphi)=\sigma(0)+\sigma'(0)\varphi+\cdots+\frac{\sigma^{(k)}(0)}{k!}\varphi^k + R_k(\varphi)
%\begin{split}
%\sigma(\varphi) = \sigma(0) +\pmb{(}d_\varphi\sigma(\varphi)|_{\varphi=0}\pmb{)} \varphi + \cdots + \pmb{(}d\varphi^{(k)}\sigma(\varphi)|_{\varphi=0}\pmb{)} \varphi^k + R_k(\varphi)\varphi^{k}.
%\end{split}
\end{equation}
where $\sigma^{(k)}$ is the $k-$order derivative of $\sigma$.

Activation functions, denoted by $\sigma$, are often chosen such that their tangent at the point $\varphi=0$ has a slope of one. This means that near the origin, the activation function behaves similarly to the identity function, remembering that $\sigma(0)=0$. Consequently, for sufficiently small values of $\varphi$, the activation function can be approximated as $\sigma(\varphi)\simeq \varphi$. This approximation holds within a regime where neuronal activity is constrained. We will refer to this as the "regular regime". It is important to clarify that the linear approximation is not exclusive to the "regular regime." Rather, within this regime, the neuronal activity is so minimal that the linear approximation is formally justified. This approximation is also applicable in scenarios involving long time periods. The rationale for this can be derived from the differential equation, where it is reasonable to assume a certain level of \textit{neuronal tranquillity} over time. By neuronal tranquillity, we mean that as time progresses, either the matrix $\pmb{A}$ (which is diagonalisable) has all its eigenvalues with negative real parts (a condition known as asymptotic stability), or the neuronal activation function, which integrates the weighted signals of each neuron, stabilises the system by flattening the overall dynamics. The linearisation of the activation function is formally defined as:

\begin{equation}\label{linealizacion}
\mathscr{L}:\pmb{\sigma}\pmb{(}\pmb{\varphi}\pmb{)}\longmapsto  \pmb{\varphi}.
\end{equation}

Applying the linearisation given by \eqref{linealizacion}, the differential equation \eqref{din general} simplifies to:

\begin{equation}\label{din general lineal}
\pmb{h}'(t)=\pmb{A}\pmb{h}(t)+\pmb{B}\,\pmb{x}(t)+\pmb{b}.
\end{equation}

Based on defined operations: rescaling $\mathscr{R}_\tau$, discretisation $\mathscr{D}_\Delta$, linearisation $\mathscr{L}$, we will study the result of applying a sequence of two operations to the dynamics regulated by \eqref{din red bias}.

Consider two computers and a scale change {$\mathscr{R}_\tau$} that stretches $\tau>1$ (compresses is given by $\tau<1$) the temporal samples. The following result can be anticipated intuitively: suppose a scale change $\mathscr{R}_\tau$ is performed and then discretised through $\mathscr{D}_\Delta$ for use with computer 1, obtaining the sequence of activities $\pmb{\mathfrak{h}}_1(s_k)$. On the other hand, if discretised with width $\Delta$ for use with computer 2 and then subsequently samples are separated (grouped) by a quantity given by $\tau$: $t_{k+1}-t_k= \tau (s_{k+1}-s_k)$, $\pmb{\mathfrak{h}}_2(s_k)$ would be obtained, and it should be verified that $\pmb{\mathfrak{h}}_1(s_k){=}\pmb{\mathfrak{h}}_2(s_k)$.  In this way, these two procedures: {$\mathscr{R}_\tau\to\mathscr{D}_\Delta$ and $\mathscr{D}_\Delta\to\mathscr{R}_\tau$ lead to the same results and are computationally indistinguishable}. We can summarise both processes discussed and the resulting modifications panel a) of Figure \ref{fig-01}.

Let us now consider the linearisation and discretisation process implemented in the equation for the activity as is shown in panel b) of Figure \ref{fig-01}. In this case, we can see how both procedures: {$\mathscr{D}_\Delta\to\mathscr{L}$ and $\mathscr{L}\to\mathscr{D}_\Delta$ lead to the same results and are also computationally indistinguishable}. 

Finally, let's consider the linearisation and rescaling procedures as shown in panel c) of Figure \ref{fig-01}. Again, we can see that the procedures {$\mathscr{L}\to\mathscr{R}_\tau$ and $\mathscr{R}_\tau\to\mathscr{L}$ are equivalent in the mentioned sense.}

\begin{figure*} % [H] fuerza la posición exacta
  %\centering
  \includegraphics[width=18cm]{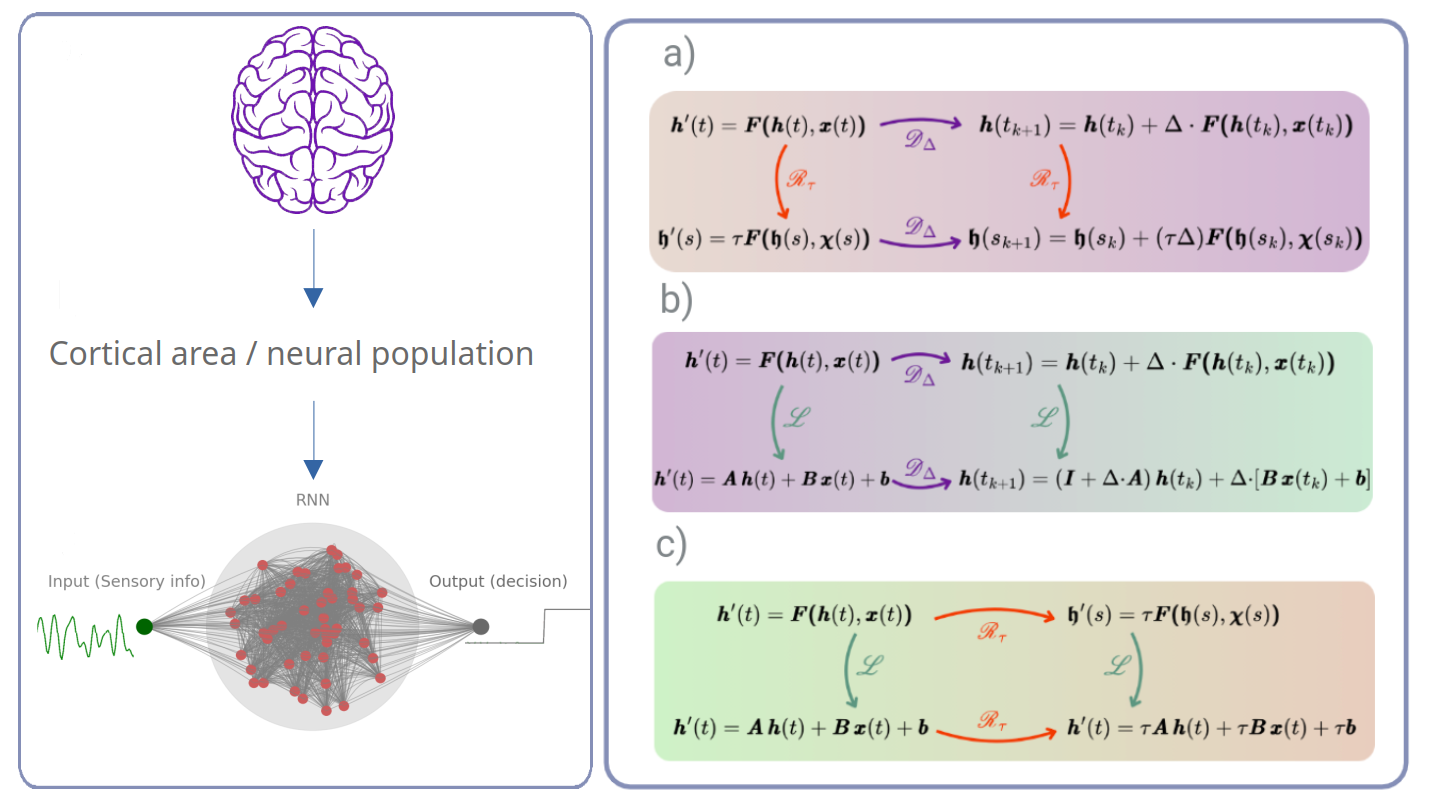} % Ajusta al ancho de una columna
  \caption{{Left panel: Schema of how an $\mathtt{RNN}$ represents a {cortical area or neural population}. Operations over the $\mathtt{RNN}$.} 
  \textbf{a)} Scheme of how the equations are modified via temporal rescaling and discretisation. 
  \textbf{b)} linearisation and discretisation process implemented in the equation for the activity.
  \textbf{c)} Linearisation and rescaling procedures implemented on the activity of units.}
  {In all panels, horizontal arrows ($\rightarrow$) represent the application of the labelled operator to the equation at the origin of the arrow, yielding the expression at the arrowhead. Vertical arrows ($\downarrow$) indicate the complementary operation applied to the alternative pathway, illustrating commutativity. Equations shown in the upper row of each panel correspond to the continuous-time formulation, while those in the lower row correspond to the transformed version.  The result at each corner of the diagram is the same regardless of the order of application, confirming that the three operations commute pairwise, in the sense that for any pair among $\mathscr{R}_\tau,\mathscr{D}_\Delta, \mathscr{L}$, their compositions are invariant under permutation.}
  \label{fig-01}
\end{figure*}

%\begin{color}{red}
\subsection{Operational equivalence of the procedures}\label{operational}

To rigorously analyse how temporal rescaling ($\mathscr{R}_\tau$), discretisation ($\mathscr{D}_\Delta$), and linearisation ($\mathscr{L}$) interact, we adopt an approach based entirely on the structural invariance of the resulting equations. Rather than formulating an abstract operator algebra over heterogeneous mathematical spaces or dealing with incompatible domains, we evaluate commutativity by directly comparing the functional form of the expressions produced by each sequence of operations.

Let $A$ and $B$ be two distinct operations chosen from the set $\{\mathscr{R}_\tau, \mathscr{D}_\Delta, \mathscr{L}\}$. If we first apply operation $A$ to the original continuous-time $\mathtt{RNN}$ equation \ref{eq-02} and subsequently apply operation $B$, we obtain a final system equation denoted as $E_{AB}$. Conversely, if we reverse the order of application, executing first $B$ and then $A$, we arrive at the final system equation denoted as $E_{BA}$.

Under this framework, we define a pair of operations to be operationally equivalent if and only if the resulting equations are analytically identical term-by-term, which we denote as:
\begin{equation}
E_{AB} \sim E_{BA}
\end{equation}
Whenever this relation holds, we can formally state that the procedures $A$ and $B$ are \textit{commutative}. 

This notion of operational equivalence formalises the diagrammatic commutativity illustrated in Figure 1. It implies that if we follow two different sequences of operations to reach a common destination in the algebraic diagram, the final functional forms and updating rules obtained via both paths coincide exactly. In the following subsections, we demonstrate analytically that this identity holds strictly for each pair of procedures, confirming that the specific order in which they are deployed does not alter the final mathematical model of the $\mathtt{RNN}$.
%\end{color}

%\begin{color}{red}

We now formally establish that the three operations introduced above commute pairwise.
Throughout, we work with the general equation \ref{eq-02}.

%-------------------------------------------------------
\subsubsection{$\mathscr{R}_\tau$ and $\mathscr{D}_\Delta$ commute}

Let $\pmb{h}(t)$ satisfy \eqref{eq-02} and let $\tau > 0$.
Applying $\mathscr{R}_\tau$ followed by $\mathscr{D}_\Delta$ yields the same
discrete sequence as applying $\mathscr{D}_{\tau\Delta}$ directly to the
original equation (equivalently, $\mathscr{D}_\Delta$ followed by a
re-indexing of step size $\tau$).

($\mathscr{R}_\tau \to \mathscr{D}_\Delta$)
Define the rescaled variable $s$ via $t = \tau s$, so that
$\pmb{h}(t) = \pmb{h}(\tau s) =: \mathfrak{h}(s)$.
Under $\mathscr{R}_\tau$, equation \eqref{eq-02} becomes
\begin{equation}
    \mathfrak{h}'(s) = \tau F\!\left(\mathfrak{h}(s), \boldsymbol{\chi}(s)\right),
    \label{eq:rescaled}
\end{equation}
where $\boldsymbol{\chi}(s) = \pmb{x}(\tau s)$.
Applying $\mathscr{D}_\Delta$ to \eqref{eq:rescaled} with the forward-Euler
approximation gives
\begin{equation}
    \mathfrak{h}(s_{k+1}) = \mathfrak{h}(s_k)
    + \tau \Delta\cdot F\!\left(\mathfrak{h}(s_k),\boldsymbol{\chi}(s_k)\right).
    \label{eq:RthenD}
\end{equation}

($\mathscr{D}_\Delta\to \mathscr{R}_\tau$) Conversely, applying $\mathscr{D}_\Delta$ first to \eqref{eq-02} with
step size $\Delta$ produces
\begin{equation}
    \pmb{h}(t_{k+1}) = \pmb{h}(t_k)
    + \Delta\cdot F\!\left(\pmb{h}(t_k), \pmb{x}(t_k)\right).
    \label{eq:DthenR}
\end{equation}
Setting $t_k = \tau s_k$, i.e., $\Delta_t = \tau \Delta_s$, and
substituting $\pmb{h}(t_k) = \mathfrak{h}(s_k)$, equation
\eqref{eq:DthenR} becomes identical to \eqref{eq:RthenD}.
Hence $\mathscr{R}_\tau$ and $\mathscr{D}_\Delta$ are commutative.

\subsubsection{$\mathscr{D}_\Delta$ and $\mathscr{L}$ commute}

Applying $\mathscr{D}_\Delta$ to the linearised equation yields the same
discrete recurrence as linearising the discretised equation.

($\mathscr{L}\to \mathscr{D}_\Delta$) 
Applying $\mathscr{L}$ (i.e., $\sigma(\varphi)\mapsto\varphi$) to
\eqref{eq-02} gives the linear system
\begin{equation}
    \pmb{h}'(t) = \pmb{A\,h}(t) + \pmb{B\,x}(t) + \pmb{b},
    \qquad \pmb{A} := \pmb{w} - \lambda I,\quad \pmb{B} := \widetilde{\pmb{w}}.
    \label{eq:linearised}
\end{equation}
Applying $\mathscr{D}_\Delta$ to \eqref{eq:linearised} yields
\begin{equation}
    \pmb{h}(t_{k+1})
    = (I + \Delta \cdot \pmb{A})\,\pmb{h}(t_k) + \Delta \cdot \pmb{B}\,\pmb{x}(t_k)
      + \Delta\cdot\pmb{b}.
    \label{eq:LthenD}
\end{equation}

($\mathscr{D}_\Delta\to \mathscr{L}$) Applying $\mathscr{D}_\Delta$ to the nonlinear equation \eqref{eq-02}
first gives
\begin{equation}
    \pmb{h}(t_{k+1}) = \pmb{h}(t_k)
    + \Delta\!\left[-\lambda\pmb{h}(t_k)
      + \sigma\!\left(\pmb{w}\pmb{h}(t_k)+\widetilde{\pmb{w}}\pmb{x}(t_k)+\pmb{b}\right)\right].
    \label{eq:DthenL_pre}
\end{equation}
Applying $\mathscr{L}$ (replacing $\sigma(\varphi)\mapsto\varphi$) to
\eqref{eq:DthenL_pre} yields
\begin{equation}
    \pmb{h}(t_{k+1})
    = \pmb{h}(t_k)
      + \Delta\!\left[-\lambda\pmb{h}(t_k)
        + \pmb{w}\pmb{h}(t_k) + \widetilde{\pmb{w}}\pmb{x}(t_k) + \pmb{b}\right]
    = (I + \Delta \cdot \pmb{A})\,\pmb{h}(t_k) + \Delta \cdot \pmb{B}\,\pmb{x}(t_k)
      + \Delta\cdot\pmb{b}.
    \label{eq:DthenL}
\end{equation}
Equations \eqref{eq:LthenD} and \eqref{eq:DthenL} are identical, so $\mathscr{D}_\Delta$ and $\mathscr{L}$ are commutative.

%-------------------------------------------------------
This result formalises and extends the observation of \citet{pagan2023brief},  who noted that linearising before versus after the pointwise nonlinearity yields equivalent dynamics; here we prove this equivalence holds exactly within the forward-Euler discretisation scheme, for the full biased continuous-time $\mathtt{RNN}$ equation \eqref{eq-02}.

\subsubsection{$\mathscr{L}$ and $\mathscr{R}_\tau$ commute}
Linearising the rescaled equation yields the same result as rescaling the linearised equation.

($\mathscr{R}_\tau\to \mathscr{L}$) 
Under $\mathscr{R}_\tau$, equation \eqref{eq-02} becomes (see
Section~2.1)
\begin{equation}
    \mathfrak{h}'(s) = -\tau\lambda\,\mathfrak{h}(s)
    + \tau\,\sigma\!\left(\pmb{w}\mathfrak{h}(s)+\widetilde{\pmb{w}}\boldsymbol{\chi}(s)+\pmb{b}\right).
    \label{eq:rescaled2}
\end{equation}
Applying $\mathscr{L}$ to \eqref{eq:rescaled2} gives
\begin{equation}
    \mathfrak{h}'(s)
    = \tau A\,\mathfrak{h}(s) + \tau B\,\boldsymbol{\chi}(s) + \tau\pmb{b}.
    \label{eq:RthenL}
\end{equation}

($\mathscr{L}\to \mathscr{R}_\tau$) Applying $\mathscr{L}$ to \eqref{eq-02} first yields
\eqref{eq:linearised}.
Applying $\mathscr{R}_\tau$ to \eqref{eq:linearised} (using the same
chain-rule argument as in Section~2.1) gives
\begin{equation}
    \mathfrak{h}'(s)
    = \tau\!\left[A\,\mathfrak{h}(s) + B\,\boldsymbol{\chi}(s) + \pmb{b}\right]
    = \tau A\,\mathfrak{h}(s) + \tau B\,\boldsymbol{\chi}(s) + \tau\pmb{b}.
    \label{eq:LthenR}
\end{equation}
Equations \eqref{eq:RthenL} and \eqref{eq:LthenR} are identical, so $\mathscr{L}$ and $\mathscr{R}_\tau$ are commutative.

The three propositions above establish the full commutativity diagram displayed in Figure~\ref{fig-01}. Every path through the
diagram yields the same equation, regardless of the order in which $\mathscr{R}_\tau$, $\mathscr{D}_\Delta$, and $\mathscr{L}$ are applied.
%\end{color}

{\subsection{Numerical verification of the commutativity properties}}

{To illustrate the analytical results of Section \ref{operational}, we implemented a small continuous-time $\mathtt{RNN}$ of $N=6$ units receiving a single smooth input $x(t)$ (a Gaussian pulse). The nonlinear activation was $\sigma(\cdot)=\tanh(\cdot)$, which satisfies $\sigma(0)=0$, $\sigma'(0)=1$, placing the network in the \textit{regular regime} under which the linearisation $\mathscr{L}$ is justified. Using the Euler discretisation $\mathscr{D}_\Delta$ with step $\Delta = 0.02$ and a temporal rescaling factor $\tau = 1.6$, we constructed the discrete recurrences corresponding to the two possible orders of each operator pair: $(\mathscr{R}_\tau,\mathscr{D}_\Delta)$, $(\mathscr{D}_\Delta,\mathscr{L})$, and $(\mathscr{L},\mathscr{R}_\tau)$.}

{For the pairs $(\mathscr{R}_\tau,\mathscr{D}_\Delta)$ and $(\mathscr{L},\mathscr{R}_\tau)$, each ordering was implemented as a separate Euler integration with its own vector field and/or step-size/time-scale schedule, so that the agreement between the two orderings is a non-trivial numerical check. For the pair $(\mathscr{D}_\Delta,\mathscr{L})$, the two orderings were implemented via two independent code paths: $\mathscr{L}\to\mathscr{D}_\Delta$ discretises the already-linearised continuous ODE \eqref{eq:linearised}, whereas $\mathscr{D}_\Delta\to\mathscr{L}$ first discretises the full nonlinear vector field with forward Euler and then applies the substitution $\sigma(\varphi)\mapsto\varphi$ to the argument of $\sigma$ within the resulting discrete update expression, without ever evaluating the linear vector field \eqref{eq:linearised} directly. Both procedures reduce analytically to the same affine recurrence (Eqs.~\eqref{eq:LthenD}--\eqref{eq:DthenL}), and their numerical agreement confirms this identity rather than merely reproducing a single computation under two names.}

{In every case, the trajectories obtained from the two orderings agreed up to floating-point precision (maximum absolute difference on the order of $10^{-16}$). Projecting the high-dimensional activity onto its first two principal components produced perfectly overlapping curves, confirming that the commutativity holds exactly under Euler discretisation and providing a concrete numerical complement to the analytic proofs (see Figure \ref{fig:2}). The code for reproducing these simulations is provided at \url{https://github.com/katejarne/Analysing_Rescaling_Discretisation_Linearisation_in_RNNs}.}

\begin{figure}
    \centering
    \includegraphics[width=0.99\linewidth]{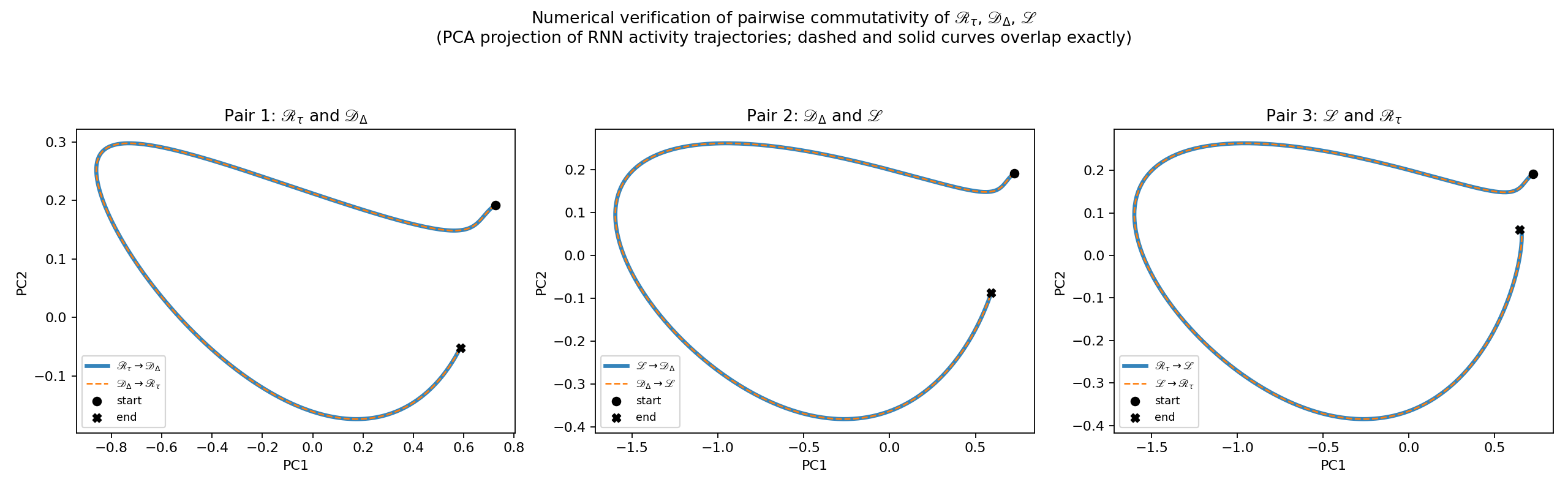}
   \caption{{Numerical illustration of the pairwise commutativity of temporal rescaling ($\mathscr{R}_\tau$), discretisation ($\mathscr{D}_\Delta$), and linearisation ($\mathscr{L}$), for a small ($N=6$) RNN of the form of Eq.~\eqref{eq-02} driven by a smooth Gaussian-pulse stimulus $x(t)$. Each
panel shows the activity trajectory $\pmb{h}$ projected onto the first two principal components (PC1, PC2) of the pooled activity, computed in the two
possible orders of application of the corresponding pair of operators; the solid blue and dashed orange curves overlap exactly, with markers indicating
the common initial (\(\bullet\)) and final (\(\times\)) states. \textbf{Panel a)} Pair $(\mathscr{R}_\tau,\mathscr{D}_\Delta)$: the
trajectory obtained by rescaling time and then discretising ($\mathscr{R}_\tau \to \mathscr{D}_\Delta$, Eq.~\eqref{eq:RthenD}) coincides
with that obtained by discretising the original equation with step $\tau\Delta$ and re-indexing the samples ($\mathscr{D}_\Delta \to \mathscr{R}_\tau$, Eq.~\eqref{eq:DthenR}); here the full nonlinear vector field $\pmb{F}$ is used in both orderings, with two genuinely different step-size/time-scale schedules.
\textbf{Panel b)} Pair $(\mathscr{D}_\Delta,\mathscr{L})$: the trajectory obtained by linearising the continuous dynamics and then discretising
($\mathscr{L} \to \mathscr{D}_\Delta$, Eq.~\eqref{eq:LthenD}) coincides with that obtained by discretising the full nonlinear equation and subsequently
linearising the resulting discrete update expression ($\mathscr{D}_\Delta \to \mathscr{L}$, Eq.~\eqref{eq:DthenL}); the two trajectories are computed via two independent numerical procedures that both reduce analytically to the same affine recurrence $\pmb{h}(t_{k+1}) = (\pmb{I}+\Delta\pmb{A})\pmb{h}(t_k)+\Delta\pmb{B}\pmb{x}(t_k)+\Delta\pmb{b}$.
\textbf{Panel c)} Pair $(\mathscr{L},\mathscr{R}_\tau)$: the trajectory obtained by rescaling and then linearising
($\mathscr{R}_\tau \to \mathscr{L}$, Eq.~\eqref{eq:RthenL}) coincides with that obtained by linearising and then rescaling
($\mathscr{L} \to \mathscr{R}_\tau$, Eq.~\eqref{eq:LthenR}); both reduce to $\mathfrak{h}'(s)=\tau\pmb{A}\mathfrak{h}(s)+\tau\pmb{B}\boldsymbol{\chi}(s)+\tau\pmb{b}$.
Note that the three panels correspond to different governing equations (nonlinear in panel a, linearised in panels b--c, and additionally
rescaled by $\tau$ in panel c), so the shapes of the trajectories differ \emph{across} panels; the relevant comparison, which confirms the
commutativity proofs of Section~2, is the exact overlap of the two curves \emph{within} each panel. In all cases, the maximum absolute difference
between the two orderings is at floating-point precision ($\sim 10^{-16}$, as reported by the accompanying script).}}
    \label{fig:2}
\end{figure}

\section{DISCUSSION}

We have delved into the fundamental mathematical framework of $\mathtt{RNN}$s and introduced procedures such as temporal rescaling, temporal discretisation, and linearisation for characterising the system's behaviour. Each of these procedures is integral to understanding and modelling $\mathtt{RNN}$s effectively.

Temporal rescaling is a crucial tool for studying $\mathtt{RNN}$s over different time scales. Under a rescaling time operation, we can observe how the dynamics of the network change, potentially uncovering information about its behaviour at different time resolutions.  This procedure is particularly useful when dealing with networks that exhibit distinct behaviours or patterns over varying time intervals. This procedure is reversible, i.e, $\mathscr{R}_\tau$ has a unique inverse given by $\mathscr{R}_{\tau^{-1}}$.

Discretisation, on the other hand, plays an important role in implementing $\mathtt{RNN}$s simulations on computers. It transforms the continuous-time differential equation into a discrete-time recurrence relation, making it computationally tractable. This is essential for simulating and analysing $\mathtt{RNN}$s in practice, allowing researchers to explore their behaviour and capabilities systematically.

Linearisation is another valuable tool that simplifies complex nonlinear $\mathtt{RNN}$s into linear approximations. This simplification aids in analysing the network's behaviour around specific operating points and designing control strategies. However, it is important to note that linearisation is most effective when nonlinearities are relatively small, and it may not be suitable for highly nonlinear systems.

Discretisation and linearisation are both irreversible procedures. However, when both procedures are applied sequentially, they yield the same result regardless of the order in which they are applied.

These procedures discussed above are standard practices in the field of $\mathtt{RNN}$s, and their order of application is often flexible, as they commute without altering the outcome. While they are powerful tools for modelling and analysing $\mathtt{RNN}$s, it's essential to choose the appropriate procedure based on the specific characteristics and goals of the neural network model being studied and compared with brain activity. {For instance, temporal rescaling might distort the alignment between model dynamics and experimentally observed neural timescales (e.g., synaptic delays or oscillation frequencies). {For example, gamma oscillations (30–80 Hz), which are tightly linked to attentional modulation and sensory processing (\cite{annurev-neuro-062111-150444}), require a discretisation step $\Delta$ small enough to resolve sub-cycle dynamics; coarse time steps can alias these oscillations or suppress them entirely, leading to qualitatively incorrect network behaviour. Similarly, synaptic delays on the order of a few milliseconds, which are physiologically critical in shaping the timing of cortical responses, may be inadvertently collapsed or distorted if that chosen $\Delta$ exceeds their duration. In such cases, the apparent dynamics of the discretised model may diverge substantially from those of the underlying continuous system, even when other qualitative features such as fixed points are preserved.}

Similarly, discretisation could introduce numerical artifacts that misrepresent continuous neural processes, such as spike timing precision. Furthermore, linearisation risks oversimplifying nonlinear phenomena critical to neural computation, like chaotic dynamics or bifurcations observed in cortical networks.} Finally, it's worth considering that these procedures may not capture the full complexity of certain networks with strong, pervasive nonlinearities or large state spaces, highlighting the need for a thoughtful approach to their application. 

%%%%%%%%%%% agregado

%\begin{color}{red}
While the pairwise commutativity of temporal rescaling, discretisation, and linearisation is an elegant mathematical property, its primary value lies in resolving operational ambiguities in everyday computational neuroscience and machine learning workflows. We highlight three prominent practical scenarios: \textbf{1.} A common paradigm in modern computational neuroscience involves "opening the black box" of trained RNNs to understand neural computation \citep{DBLP:journals/neco/SussilloB13, maheswaranathan2019reverse}. Because training via backpropagation through time (BPTT) requires discrete steps, researchers inherently obtain a discrete-time model, which they subsequently linearise to find fixed points and compute Jacobians. A persistent theoretical ambiguity is whether the stability properties (eigenvalues) derived from this discrete approximation faithfully represent the underlying continuous biological system. The exact commutativity between $\mathscr{L}$ and $\mathscr{D}_\Delta$ proven in this work acts as a formal guarantee: linearising the trained discrete model yields the identical dynamical landscape as linearising the original continuous ODE and subsequently discretising it. Consequently, researchers can confidently infer continuous biological attractors directly from discrete \textit{in silico} Jacobians. \textbf{2.} In studying cognitive processes such as motor timing \citep{Wang2018}, modellers often need to observe how a neural circuit behaves at different temporal speeds. If an RNN has already been trained with a specific integration step $\Delta$ and time constant $\tau$, scaling the network's speed poses a methodological choice: should one scale the continuous equations and re-discretise (potentially requiring retraining), or simply manipulate the recurrent matrix of the already discrete model? The commutativity between $\mathscr{D}_\Delta$ and $\mathscr{R}_\tau$ ensures that both approaches are analytically equivalent. Modellers can directly scale the discrete equations post-training without losing fidelity to the continuous system, thereby saving substantial computational resources and preventing numerical instability. \textbf{3.} A standard procedure consists of characterising the behaviour of the system near its equilibrium points by computing the Jacobian matrix (linearisation $\mathscr{L}$). When studying how these circuits adapt their processing speed to external demands through a temporal scale factor (rescaling $\mathscr{R}_\tau$), an ambiguity arises as to whether the local linear approximation is distorted or whether the nature of the attractors changes depending on the order of the analysis. The exact commutativity between $\mathscr{R}_\tau$ and $\mathscr{L}$ provides a fundamental guarantee of invariance: linearising the temporally modified vector field produces exactly the same local phase portrait as temporally rescaling the original linearised system. The resulting eigenvalues are simply multiplied homogeneously by the factor $\tau$, strictly preserving the signs, the directions of the eigenvectors, and the topological classification of the fixed point (whether it is a node, a focus, or a saddle point). This ensures that the computational and qualitative properties of the network do not suffer from geometric artefacts when altering the intrinsic speed of the system.
%\end{color}

%%%%%%%%%%%

{Beyond the neuroscience applications discussed here, the broader challenge of processing and protecting temporally and spatially structured medical data has motivated parallel advances in clinical deep learning. Recurrent and sequence-aware architectures are involved in multimodal medical imaging pipelines, including systems that must track patient identity across longitudinal studies while preserving clinically relevant features. Recent work has demonstrated that deep neural networks can simultaneously handle patient re-identification across diverse imaging modalities (\cite{Tian2025MedReID}) and protect sensitive identity information while retaining diagnostically critical signs in ophthalmic imagery (\cite{Tian2025ROFI}). These developments illustrate how the mathematical principles explored here — including stable temporal dynamics and appropriate discretisation — are increasingly relevant not only for neural modelling but also for the design of trustworthy clinical AI systems that operate on sequential or multi-session medical data.}
%%%%

% agregada
{Crucially, the pairwise commutativity of these operations guarantees the structural invariance of the network's state-space reachability, also known as controllability \cite{Sontag1997, Sontag1998}. In realistic biological models where the input dimension is strictly less than the number of neurons ($M \le N$), the network achieves full state-space exploration not in a single computational step, but through temporal integration. By preserving the algebraic independence of $\mathscr{R}_\tau$, $\mathscr{D}_\Delta$, and $\mathscr{L}$, the commutativity ensures that the system's controllability matrix maintains its full rank across domains. Consequently, if a continuous, non-linear $\mathtt{RNN}$ can theoretically be driven to an arbitrary target state via its external inputs, this topological property is faithfully preserved in its discretised, rescaled, and linearised computational counterpart \cite{Sontag1997}. This mathematical guarantee precludes the emergence of artificial "blind spots" or inaccessible sub-spaces in the simulated neural dynamics, regardless of the chosen integration step or temporal scale.}

%%%%%
{The commutativity results established here have direct practical implications for researchers building and training $\mathtt{RNN}$  models of neural dynamics. First, they justify the common but rarely formalised practice of interchanging the order of operations during model construction: for instance, a practitioner may choose to linearise an $\mathtt{RNN}$ analytically before implementing it in discrete time, or conversely, to implement the full nonlinear system and linearise the resulting map, obtaining equivalent results in either case. Second, in training paradigms that involve temporal rescaling to match experimental recordings at different timescales (e.g., interval timing tasks as in \cite{Wang2018}), the commutativity with discretisation guarantees that rescaling the continuous model before choosing a time step is equivalent to adjusting the step size after the fact, removing an otherwise ambiguous modelling decision. Third, commutativity with linearisation ensures that fixed-point stability 
conclusions are consistent across the two common analytical pathways:  whether one linearises the continuous system and then discretises the 
resulting linear map, or discretises the nonlinear system first and then linearises the discrete map around a fixed point, the resulting recurrence matrix $(I + \Delta\cdot \pmb{A})$ is the same in both cases.
}

%%%%%

While these techniques offer powerful tools for understanding and modelling neural processes, {future work could investigate whether analogous commutativity properties persist under higher-order numerical integrators, adaptive-step solvers, or linearisations performed around non-trivial attractors rather than near the origin.}

\section*{Conflict of Interest Statement}
The authors declare that the research was conducted in the absence of any commercial or financial relationships that could be construed as a potential conflict of interest.

\section*{Author Contributions}
M.C. and C.J. designed the study. MC performed analytical derivations. C.J. supervised and contributed to the interpretation. Both authors wrote and approved the manuscript.

\section*{Funding}
C.J. is supported by UNQ project 1520/25 and CONICET.  M. C. acknowledges support from UNIR and FIDESOL-related resources.

\section*{Acknowledgments}
We thank colleagues and collaborators for helpful discussions on continuous-time $\mathtt{RNN}$ modelling and dynamical systems.

\section*{Data Availability Statement}
No datasets were generated or analysed in this theoretical study.

\bibliographystyle{Frontiers-Harvard} 
\bibliography{references}

\end{document}